\documentclass[sigconf]{acmart}

\renewcommand\footnotetextcopyrightpermission[1]{} 
\usepackage{booktabs} 
\usepackage[linesnumbered,boxed,ruled]{algorithm2e}
\usepackage{graphicx, subfigure}
\usepackage{float}
\usepackage{enumitem}

\acmConference[SMASC'17@CIKM]{ACM conference}{Nov. 2017}{Singapore} 

\begin{document}
\title{Time-sensitive Customer Churn Prediction based on PU Learning}

\author{Li Wang, Chaochao Chen, Jun Zhou, Xiaolong Li}
\affiliation{%
  \institution{Ant Financial Group}
  \streetaddress{\{raymond.wangl,chaochao.ccc,jou.zhoujun,xl.li\}@antfin.com}
}

\begin{abstract}
With the fast development of Internet companies throughout the world, customer churn has become a serious concern. 
To better help the companies retain their customers, it is important to build a customer churn prediction model to identify the customers who are most likely to churn ahead of time. 
In this paper, we propose a Time-sensitive Customer Churn Prediction (TCCP) framework based on Positive and Unlabeled (PU) learning technique. 
Specifically, we obtain the recent data by shortening the observation period, and start to train model as long as enough positive samples are collected, ignoring the absence of the negative examples. We conduct thoroughly experiments on real industry data from Alipay.com. The experimental results demonstrate that TCCP outperforms the rule-based models and the traditional supervised learning models.
\end{abstract}

%


\maketitle

\section{Introduction}

The newly raising Internet companies are facing the same problem that bothers many traditional industries, i.e., customer churn. 
A common way to retain customers is to provide incentives for those who are most likely to churn. 

Supervised learning are studied to solve the churn prediction problem. 
To do this, both positive and negative samples are needed. 
Commonly, customers who have stopped using a company's services for a period of time are defined as negative (churned) samples, i.e., $y=0$, and others as positive samples, i.e., $y=1$. 
Methods like decision tree \cite{wei2002turning}, SVM \cite{coussement2008churn}, neural network \cite{sharma2013neural,tsai2009customer}, and random forest \cite{xie2009customer} are applied to train the classifier. 
However, it always takes months to determine a candidate customer would become a negative sample or not. 
However, the Internet companies are developing rapidly, and 
the classifier trained on the data of months ago would be outdated. 
How to obtain a customer churn prediction model with better timeliness 
becomes a hot research area. 

\textbf{P}ositive and \textbf{U}nlabeled (\textbf{PU}) learning is studied to deal with the situations where only positive and unlabeled samples are available \cite{liu2002partially,li2003learning}. 
Among these methods, \textbf{weighting sample} approaches are well studied, e.g., weighted logistic regression \cite{lee2003learning} and weighted SVM \cite{elkan2008learning}. 
They consider unlabeled data as negative samples and weight positive and negative samples to train a classifier. 
PU learning technique seems to be an ideal way for customer churn prediction problem, since one can train a classifier when enough positive samples are collected and ignore the absence of negative (churned) data. 
Surprisingly, no similar techniques have been studied before.

In this paper, 
we take the customers who use the company's services in a certain time period as positive samples, and the others as the unlabeled samples, since we do not know if they will come back in next few months yet.
Our main contributions are summarized as follows: 
(1) We propose a novel idea, i.e., applying PU learning technique to solve the customer churn prediction problem. 
(2) We propose a  \textbf{Time-sensitive Customer Churn Prediction (TCCP)} framework, which is a novel realization of PU learning technique in churn prediction scenarios. 
(3) We perform throughly experiments on real industry dataset, and the results demonstrate that TCCP achieves better prediction performance comparing with the rule-based methods and the traditional supervised learning methods.

\section{Preliminary}

\subsection{PU Learning by Weighting Samples}
Let  $x$ represent a sample,  $y\in \{0, 1\}$ represent its label, and  $s\in\{0,1\}$ represent whether the sample is labeled. 
In PU learning scenario, only positive samples are labeled, i.e., $p(s=1|x,y=0)=0$, which indicates no negative sample is labeled.


The goal is to learn a classifier $f(x)$, such that $f(x) = p(y=1 | x)$ as close as possible. 
We also assume that the labeled positive samples are chosen completely randomly from all the positive samples, i.e., $p(s=1|x,y=1)=p(s=1|y=1)$, and then we have
\begin{equation}\label{eq1}
  p(y = 1|x) = p(s = 1|x)/c,
\end{equation}
where$c = p(s = 1|y = 1)$.

Clearly \begin{math}p(y = 1|x, s = 1) = 1\end{math}, because all the labeled samples are positive.  Based on Eq.(\ref{eq1}), we have
\begin{equation}\label{eq11}
\begin{aligned}
p(y = 1|x, s = 0) & = \frac{p(s=0|x,y=1)p(y=1|x)}{p(s=0|x)} \\
		&	= \frac{[1-p(s=1|x,y=1)]p(y=1|x)}{1-p(s=1|x)} \\					
		&	= \frac{(1-c)p(y=1|x)}{1-p(s=1)|x}		\\				
		&	= \frac{1-c}{c} \frac{p(s=1|x)}{1-p(s=1|x)}. \\
\end{aligned}
\end{equation}





We now describe a classic PU learning technique, i.e., weighting samples \cite{elkan2008learning}, since we apply it into churn prediction problem. 
We view $x$, $y$, and $s$ as random variables, and the goal is to estimate \begin{math}{E}_{p(x,y,z)}[h(x,y)]\end{math} for any function $h$, where $p(x, y, s)$ is the overall distribution over triples $<x, y, s>$. We want to estimate $E[h]$ based on $<x,s>$, and by definition:

\begin{math}
\begin{aligned}
E[h]  &= \int_{x,y,s}h(x,y)p(x,y,s)\\
	&= \int_{x}p(x)\sum_{s=0}^{1}p(s|x)\sum_{y=0}^{1}p(y|x,s)h(x,y) \\
	&= \int_{x}p(x)(p(s=1|x)h(x,1) \\
	&	\qquad\qquad		+p(s=0|x)[p(y=1|x,s=0)h(x,1) \\
	&	\qquad\qquad\qquad	+p(y=1|x,s=0)h(x,0)]).\\
\end{aligned}
\end{math}

The plugin estimate of $E[h]$ is the empirical average:
\begin{equation}\label{eq2}
\frac{1}{m}(\sum_{<x,s=1>}^{}h(x,1)+\sum_{<x,s=0>}^{}w(x)h(x,1)+(1-w(x))h(x,0)),
\end{equation}
where $w(x)=p(y=1|x,s=0)$
and $m$ is the cardinality of the training set. 
Based on Eq.(\ref{eq11}), $w(x)=\frac{1-c}{c}\frac{g(x)}{1-g(x)}$, where $g(x)=p(s=1|x)$. 

From Eq.(\ref{eq2}), we find that each unlabeled sample can be viewed as a weighted positive sample as well as a weighted negative sample. 
This idea enables us the following implementation. First, give all the positive samples unit weight; Second, duplicate unlabeled samples, one as positive samples with weight $w(x)$, and the other as negative with weight $1-w(x)$; At last, apply supervised learning algorithm on the above samples with individual weights to obtain a classifier. 
We will describe how to apply weighting sample based PU learning technique in churn prediction scenarios in Section 3.3.



\subsection{Distributed Factorization Machines}
Factorization machines (FM) \cite{rendle2010factorization} is a classic high-order prediction model, which is a combination of a linear model and several nonlinear models. 
Take the second-order FM for sample, the linear part is used to capture the linear combinations of different single features and the nonlinear one aims to learn the weights of the interactions between different features pairs. 
Assume $\textbf{x} \in {\mathbb{R}}^{p}$ are the real valued input vectors, and then the second-order FM model can be formulated as 
\begin{equation}
\hat{y}(x):={w}_{0} + \sum_{j=1}^{p}{w}_{j}{x}_{j} + \sum_{j=1}^{p}\sum_{j'=j+1}^{p}{x}_{j}{x}_{j'}\sum_{f=1}^{k}{v}_{j,f}{v}_{j',f},
\end{equation}
where $k$ is the dimensionality of the factorization, $ \textbf{w}$ is the weight for linear features, and $\textbf{V}$ is the weight for features interactions.
Thus, the model parameters $\Theta  = \{{w}_{0}, {w}_{1}, . . . , {w}_{p}, {v}_{1,1}, . . . {v}_{p,k}\}$ are
\begin{displaymath}
{w}_{0}\in\mathbb{R}, \textbf{w}\in{\mathbb{R}}^{p},\textbf{V}\in{\mathbb{R}}_{p\times k}.
\end{displaymath}



FM offers good performance and useful embeddings of data. 
However, the basic FM is difficult to scale to large amounts of data and large numbers of features. 
Thus, a distributed Factorization Machine is proposed in \cite{li2016difacto}, which uses a refined FM model with sparse memory adaptive constraints
and frequency adaptive regularization. It allows us to distribute FM over multiple machines using the Parameter Server framework by computing distributed sub-gradients on mini-batches asynchronously. 
To improve the scalability of our proposed framework, we apply the distributed FM as the final classifier, i.e., $f(x)$. 

\section{Churn prediction Framework based on PU learning}

In this section, first we will give customer churn a formal definition from the company's perspective. Then we will introduce the motivation of our work. At last, we will introduce the proposed TCCP framework in details. 

\subsection{Problem Definition}
From the customers' position, churn is a simple action that switch one company's service to another. While the companies do not have the information that their customers have churned. All they know is that the customer has not using their service for a certain time. 
Thus, the definition of customer churn from the company's perspective is as: 
if a customer do not login a company's APP for a time period, i.e., \textbf{Churn Period (CP)}, he/she becomes a churned customer. 

Commonly, $CP$ is like 3 to 6 months for the Internet companies like \textsl{Alipay.com}. 
Our problem is how to obtain a classifier based on more recent user behaviors despite of a large \textsl{CP}.

\subsection{Motivation}
Let ${S}_{t}$ represent the set of customers who are still active at time $t$ in an APP. We call ${S}_{t}$ candidate samples at time $t$. 
We observe the customers in ${S}_{t}$ to see if they would login the APP for a period time which we term as \textbf{Observation Period (OP)}. 

The traditional supervised learning models need both positive and negative samples. 
To apply them into customer churn prediction problem, we label the customers who login the APP during the $OP$ as positive samples, and others as negative ones. 
Therefore, we need to observe a time period that longer than $CP$ to make sure a candidate sample become a positive sample or a negative one, i.e., $OP >= CP$.

Let ${t}_{1}$ represent the time of $OP$ starts, and ${S}_{{t}_{1}}$ denote the training sample set. 
Let ${P}_{{t}_{1}}$ represent the positive samples, and ${N}_{{t}_{1}}$ represent the negative ones. 
Obviously, ${P}_{{t}_{1}}$ and ${N}_{{t}_{1}}$ are both subsets of ${S}_{{t}_{1}}$.  A traditional classifier ${f}_{{t}_{1}}(x)$ can be obtained by using data ${P}_{{t}_{1}}$ and ${N}_{{t}_{1}}$. 
Obviously, the traditional classifier ${f}_{{t}_{1}}(x)$ is based on the data before time ${t}_{1}$, which is at least $CP$ time before the classifier is used. 

Internet companies like \textsl{Alipay.com} are developing rapidly nowadays, and $CP$ (usually 3 to 6 months) is such a long time  that the users' behavior patterns may be totally different. This will cause the trained classifier seriously outdated. 
Thus, using more recent training data by shortening $OP$ is essential to improve the customer churn prediction performance.

\begin{figure}[t]
\centering
\includegraphics[width=8cm]{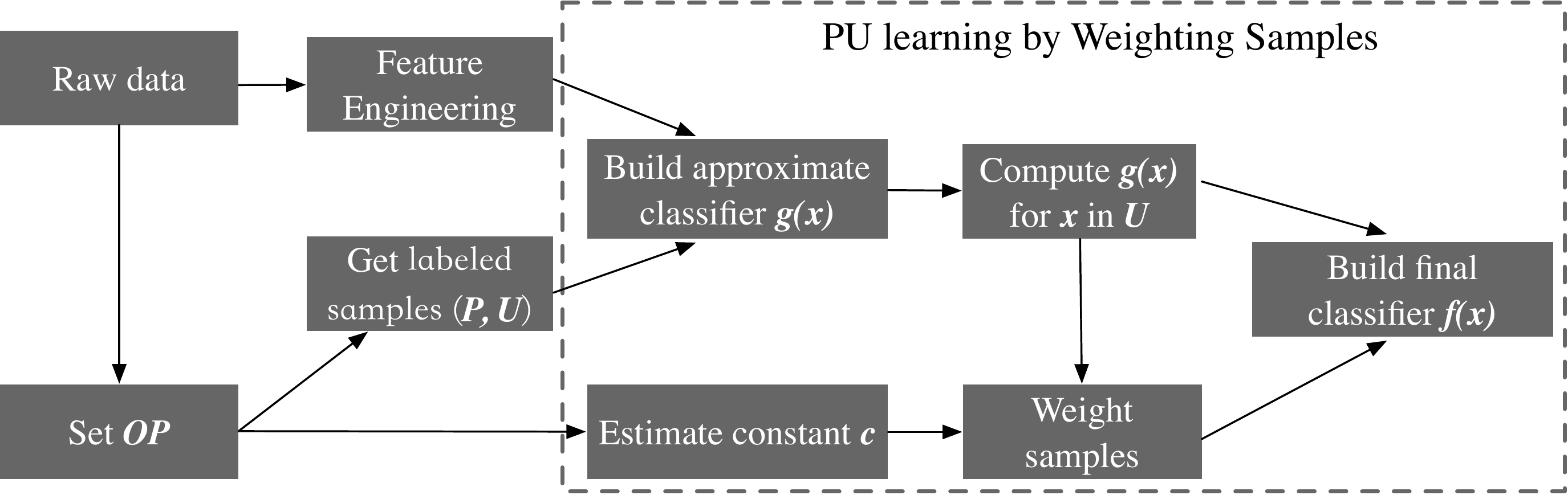}
\vskip -0.05in
\caption{TCCP framework overview.}
\label{TCCPframework}
\vskip -0.15in
\end{figure}

\subsection{The Proposed TCCP Framework}\label{framework}
We now present the proposed Time-sensitive Customer Churn Prediction (TCCP) framework. The basic idea of our framework is that we do not need to obtain the negative samples. 
We form our training data from two parts, one is the existing positive samples, and the other is the rest unlabeled samples. 
We then apply PU learning technique based on these training data. 
The TCCP framework is shown in Figure ~\ref{TCCPframework}, which generally includes the following procedures.

\subsubsection{Get labeled samples.}
Unlike the traditional supervised learning methods that require $OP >= CP$, we propose to choose a short $OP$ so that $OP < CP$. 
Let  ${t}_{2}$ represent the new start time of $OP$, and ${S}_{{t}_{2}}$ denote the new training sample set. 
Then, we label the customers in ${S}_{{t}_{2}}$ who login the APP during the $OP$ as positive samples (${P}_{{t}_{2}}$). 
For those who do not, we cannot decide whether they are negative samples or not, since $OP < CP$. 
Thus, they are unlabeled samples (${U}_{{t}_{2}}$). 
After we get positive and unlabeled training samples (${P}_{{t}_{2}}$ and ${U}_{{t}_{2}}$), we can apply PU learning technique to train a classifier ${f}_{{t}_{2}}(x)$. Because of the more recent data we use, our method can capture user behavior patterns more precisely. 
Therefore, our method can achieve better results, as will be shown in experiments.

\subsubsection{Feature Engineering.}
Let ${X}_{t}$ represent the features of training samples at time $t$, which are combined by three parts. 
\begin{itemize}[leftmargin=*] \setlength{\itemsep}{-\itemsep}
\item Customers' static profiles, like age, gender, city, register time, user level, etc.
\item Customers' dynamic preferences extracted from their behavior data. 
We generate these features by counting users' behaviors in a certain time period before time $t$, e.g., login times in the last 7 days, pay amount in the last 15 days, click counts on some page in the last 30 days, etc. 
\item Cross features that are generated from the first and the second parts of features. 
For example, the cross of age and login times can capture the relationship between `age'-`login times' pairs and churn tendency.
\end{itemize}

\subsubsection{PU Learning by Weighting Samples.}
We now describe how to apply weighting sample based PU learning technique in custom churn prediction scenarios. 
It mainly consists of the following steps. 


\begin{itemize}[leftmargin=*] \setlength{\itemsep}{-\itemsep}
\item  \textbf{Build the approximate classifier $\textbf{g(x)}$}. 
The purpose of the classifier $g(x)$ is to restrict $g(x) \approx p(s=1 | x)$ which ranges in $[0,c]$. 
To do this, we first apply Logistic Regression (LR) to train a classifier $g'(x)$ based on ${P}_{t}$ and ${U}_{t}$, and thus $g'(x) \in [0,1]$. 
We then get $g(x)$ by mapping $g'(x)$ to $g(x)$ through $g(x)=cg'(x)$. 
\item \textbf{Estimate constant $\textbf{c}$}. 
Note that $c$ denotes the probability of a positive sample being labeled, that is, $c = p(s = 1|y = 1)$. 
Thus, $c$ can be estimated by using existing statistics in custom churn prediction scenarios. 
Specifically, let ${o}_{t}$ be the observation period at time t, and ${o'}_{t}$ be the $OP$ that have ${o'}_{t} = CP$.   
Let ${P}_{{o}_{t}}$ be the positive sample set in ${o}_{t}$, which is the labeled positive samples, and ${P}_{{o'}_{t}}$ be the positive sample set in ${o'}_{t}$, which is the actual positive samples. By the definition of $c$, we have $c \approx |{P}_{{o}_{t}}|/|{P}_{{o'}_{t}}|$. 
\item \textbf{Weight samples}. 
We apply the PU learning technique described in Section 2.1 to weight samples. 
Specifically, we first give all the positive samples unit weight, i.e. $w(x)=1$. 
We then duplicate unlabeled samples. One copy is viewed as the positive samples, and are given weight $w(x) = \frac{(1-c)*g(x)}{c*(1-g(x))}$; The other copy is viewed as the negative samples, and are given weight $w(x) = 1 - \frac{(1-c)*g(x)}{c*(1-g(x))}$.
\item \textbf{Build final classifier $\textbf{f(x)}$}. We apply distributed Factorization Machine model \cite{li2016difacto} to obtain the final classifier $f(x)$, for its good ability to scale to big data.
\end{itemize}

We summarize the implementation details of PU learning by weighting samples in Algorithm 1. 
\vskip -0.1in
\begin{algorithm}[t]
\caption{Implement PU Learning by Weighting Samples.}\label{learning}
\KwIn {Positive samples ($P$), Unlabeled samples ($U$), Estimated constant ($c$) }
\KwOut{Classifier $f(x)$}
Apply LR on P and U, obtain nontraditional classifier $g'(x)$;\\
Let $g(x) = cg'(x);$\\
${W}_{P}=[]$;
${W}_{N}=[]$;
$N = []$;\\
\For{$x$ in $P$} {
	${W}_{P}$.append(1);\\
}
\For{$x$ in $U$}{
	${w}_{P}=\frac{1-c}{c}\frac{g(x)}{1-g(x)}$;\\
	${w}_{N}=1-{w}_{P}$;\\
	$P$.append($x$);\\
	${W}_{P}$.append(${w}_{P}$);\\
	$N$.append($x$);\\
	${W}_{N}$.append(${w}_{N}$);\\
}
Apply FM on data ${{W}_{P}}^{T}P$ and ${{W}_{N}}^{T}N$, obtain classifier $f(x)$.

\Return $f(x)$
\end{algorithm}
\vskip -0.15in

\section{Experiment}

In this section we report comprehensive experiments to show the effectiveness of our proposed TCCP framework on real industry data. We use both rule-based methods and traditional supervised learning method as baselines. We also study the effects of different \textsl{OP} on model performances.

\subsection{Datasets}

All our experiments are based on the real industry data from \textsl{Alipay.com}, which is the world's leading third-party payment platform.  
We choose its active customers as candidate samples at the time of OP starts. We define active customers as those who login Alipay in the past 30 days. 
Usually, most of the candidates will return to Alipay, and thus the samples are seriously imbalanced. 
To reduce the positive samples, we filter out the customers who login the APP more than 12 times in the past 30 days. 
Because those customers are rarely going to churn according to our statistics.  

We vary \textsl{OP} in $\{3, 7, 15, 30, 60, 90\}$ to get different datasets that include positive ($P$), negative ($N$), and unlabeled ($U$. 
In special case, when $OP >= CP$, the dataset does not have $U$ any more, since we can get $P$ and $N$. 
The description of training and test datasets are listed in Table~\ref{tab:datasets}, each of which consists of about 200 millions customers. 
For example, \textsl{PU-3} means \textsl{OP=3}, and similar meanings for other notations.

\begin{table}\small
\vskip -0.05in
  \caption{Datasets illustration.}
  \vskip -0.05in
  \label{tab:datasets}
  \begin{tabular}{ccccccc}
    \toprule
    Name&Period& $OP$ (day)&Usage & $P$ & $N$ & $U$\\
    \midrule
    \textbf{PU-3}&20160729-20160801&3&train &\checkmark&&\checkmark\\
    \textbf{PU-7}&20160725-20160801&7&train &\checkmark&&\checkmark\\
    \textbf{PU-15}&20160717-20160801&15&train &\checkmark&&\checkmark\\
    \textbf{PU-30}&20160703-20160801&30&train &\checkmark&&\checkmark\\
    \textbf{PU-60}&20160603-20160801&60&train &\checkmark&&\checkmark\\
    \textbf{PN-90}&20160504-20160801&90&train &\checkmark&\checkmark&\\
    \textbf{PN-90-T}&20160801-20161029&90&test &\checkmark&\checkmark&\\
  \bottomrule
\end{tabular}
\end{table}

\begin{figure}[t]
\vskip -0.05in
\centering
\subfigure [Recency rule.]{\includegraphics[width=2.8cm]{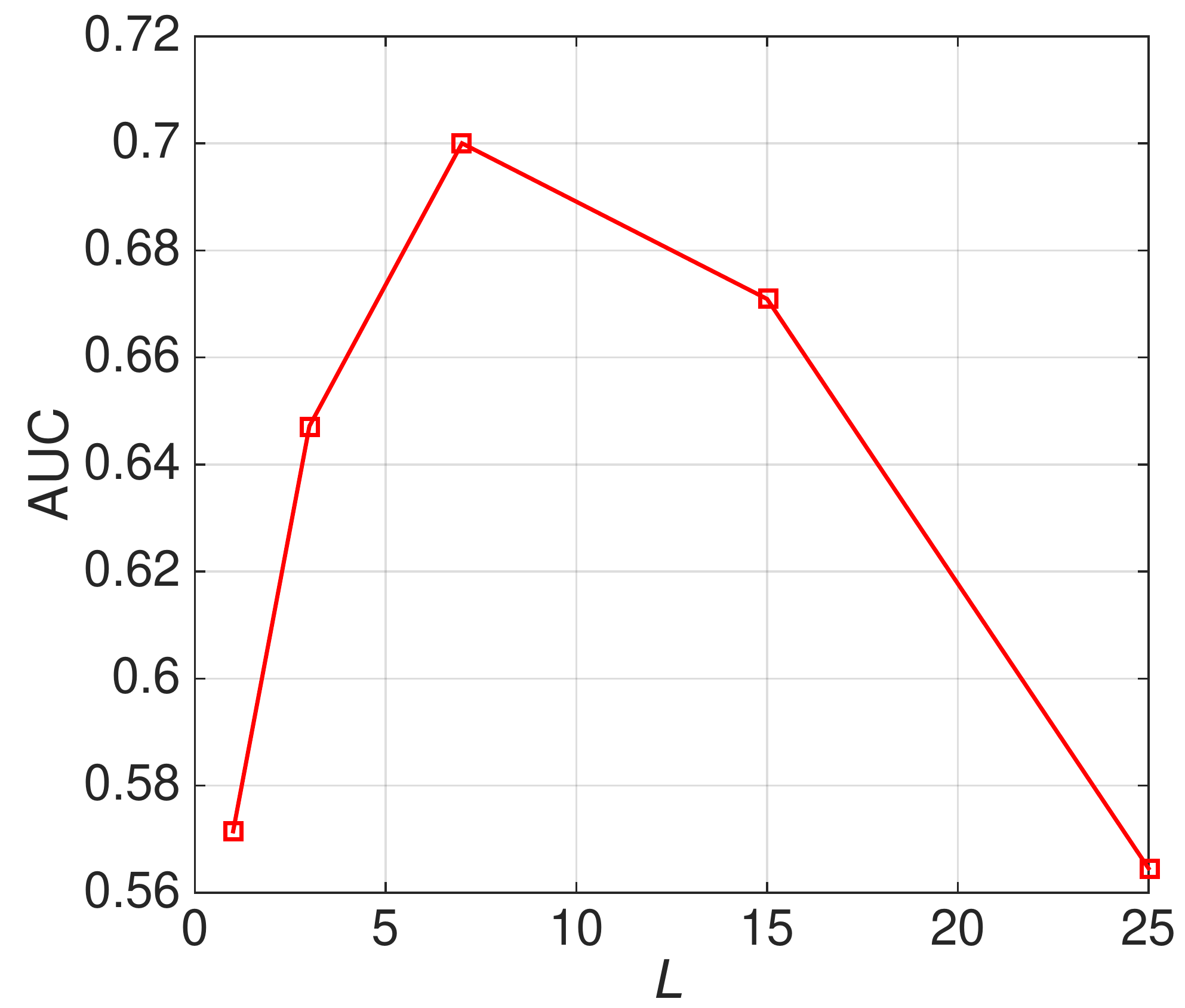}}
\subfigure [Frequency rule.] {\includegraphics[width=2.65cm]{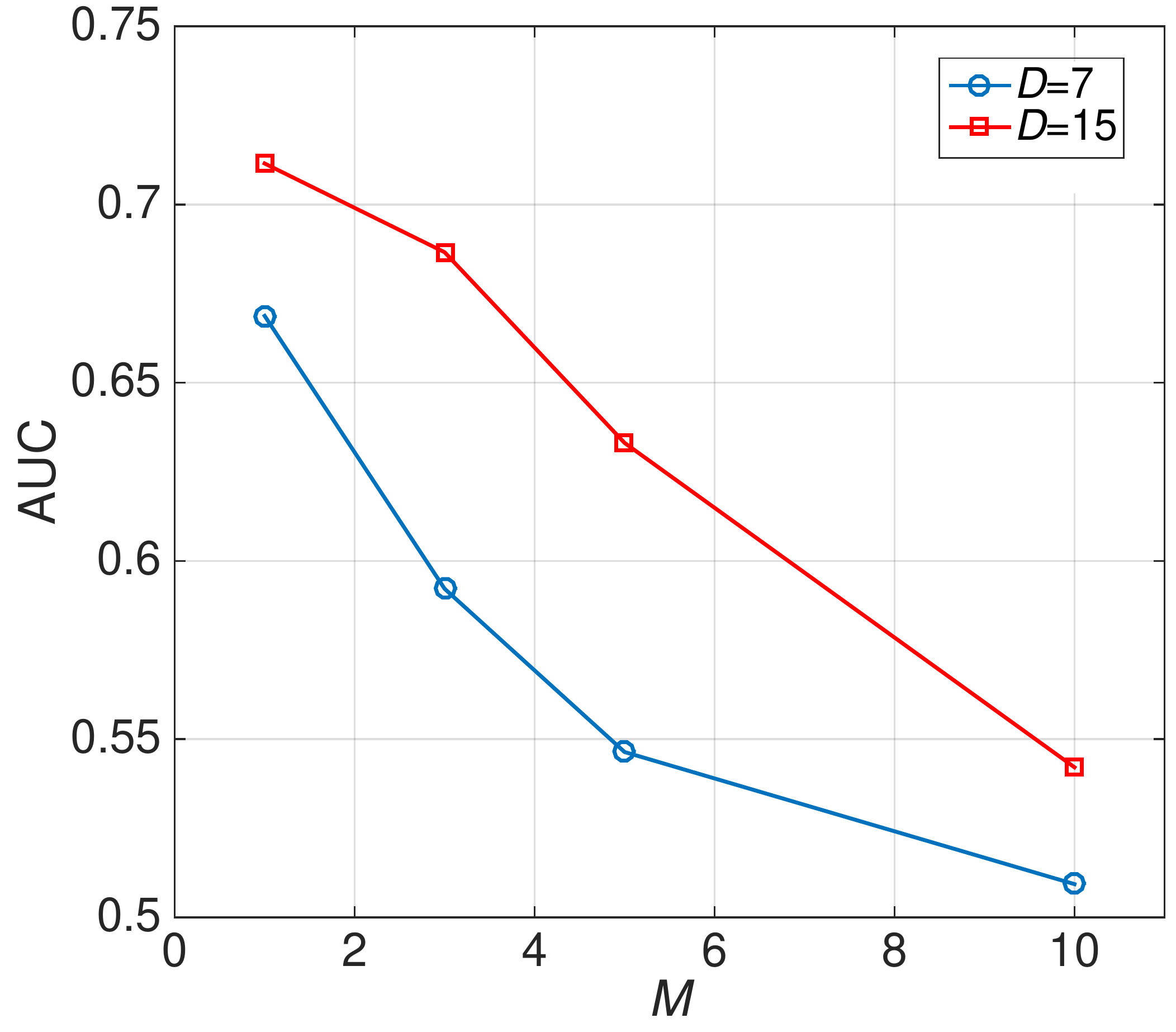}}
\subfigure [TCCP.] {\includegraphics[width=2.81cm]{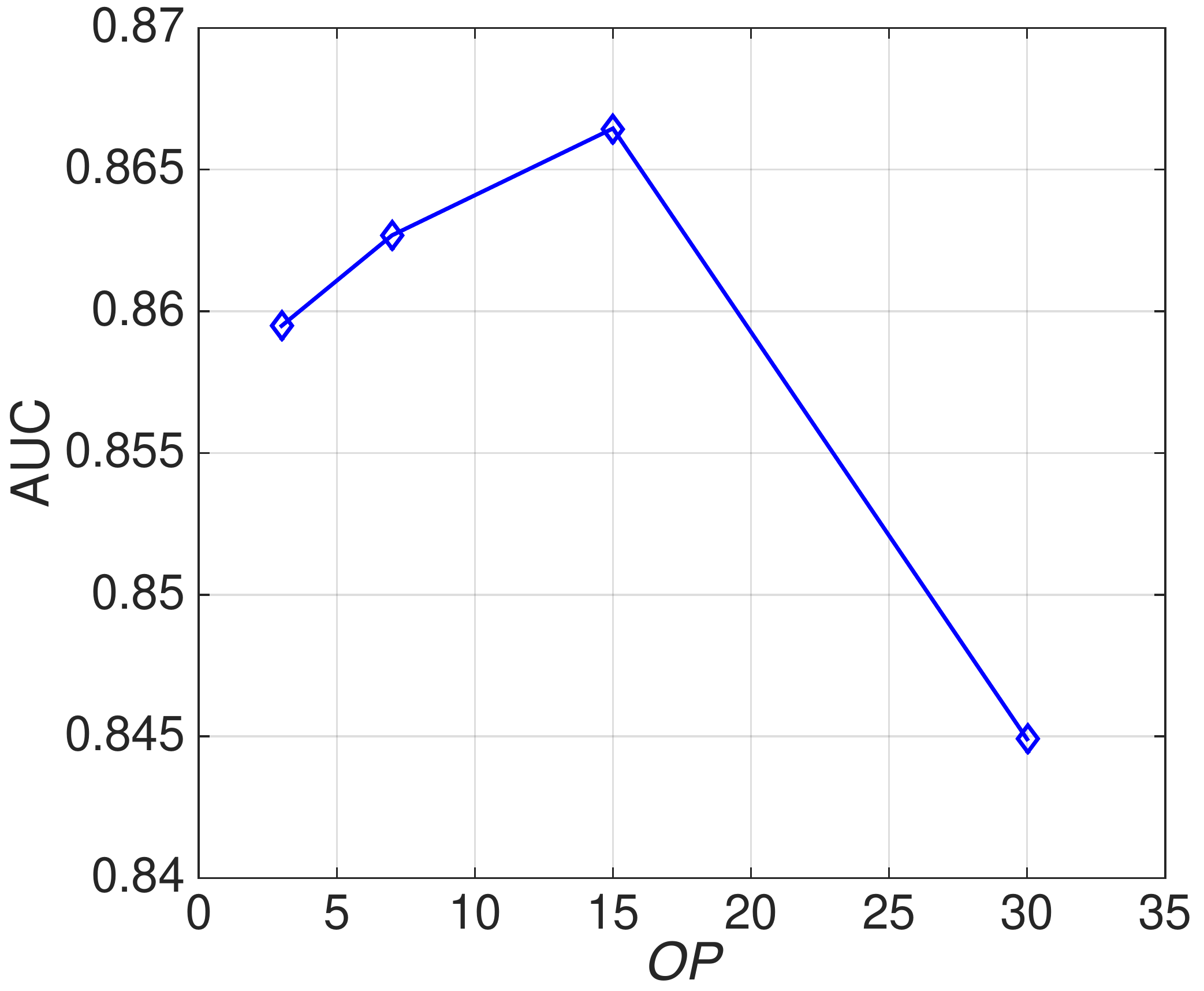}}
  \vskip -0.05in
\caption{Experiments results.}
\vskip -0.02in
\label{rule-base-result}
\end{figure}

\subsection{Baselines}

We use both rule-based methods and traditional supervised learning methods as baseline methods in our experiments. We test all classifiers on dataset \textbf{PN-90-T}, and use \textbf{Area Under the Curve (AUC)} as metric. 

\textbf{Rule based methods.}
We pick the rules from two perspectives: 
(1) Recency rule: If a customer have not login the APP for $L$ days, he/she will churn in the next 3 months. We choose different $L$ to compare the results. .
(2)Frequency rule:  If a customer login the APP less than $M$ times in the past $D$ days, he/she will churn in the next 3 months. We also choose different $D$ and $M$ to compare the results. 

\textbf{Traditional supervised learning method.} 
We choose Logistic Regression and Distributed Factorization Machine as two traditional classification methods. The reason is that they both have good ability to scale to big data.

\subsection{Comparison Results}
We report the comparison results of all the methods in Table ~\ref{compare-all}, where we set the parameters of each method to their best values. 
As we can see, two rule-based methods perform the worst, followed by two traditional supervised learning methods. Our TCCP framework performs the best. 
Specifically, it increases the AUC by 23.7\% and 21.8\% against two rule-based methods respectively, and 2.6\% and 2.4\% against two traditional supervised learning methods respectively. 
This is because our proposed TCCP framework benefits from the more recent training data and makes up the incomplete labels by applying PU learning technique. 

\subsection{Parameter Analysis}
Figure ~\ref{rule-base-result}(a) shows the results of recency rule based method. From it, we find that the prediction performance first increases with $L$, and then starts to decrease after a certain threshold. This is because a too small value of $L$ will miss-predict active customers as churn ones, while a big value of $L$ can not identify the churn customers.  
Figure ~\ref{rule-base-result}(b) shows the results of frequency rule based method. From it, we find that the prediction performances decrease with the increase of $M$. This is intuitive, since the less a customer logins the APP, the more likely he/she will churn.

Figure ~\ref{rule-base-result}(c) shows the results of our proposed TCCP framework. 
From it, we can see that the prediction performance first increases with $OP$, and starts to decrease after a certain threshold. 
This is because, TCCP uses more recent training data by sacrificing the completeness of the labels. 
When $OP$ is too small, there will be too few labeled positive samples to obtain a fine classifier, despite of the recent data. 
When $OP$ is too big, the number of labeled positive samples increases, however, the recency of the data is sacrificed. 
That is, TCCP uses out-dated data to train the model when $OP$ is too big. 
This is why the AUC performance of TCCP (0.845 when $OP=30$) becomes similar with the AUCs of two traditional supervised learning methods (0.844 and 0.846 respectively) when $OP$ increases.

\begin{table}[t]\small
	\vskip -0.05in
  \caption{Comparison results for each model.}
  \vskip -0.15in
  \label{compare-all}
  \begin{tabular}{ccc}
    \toprule
    Method& Parameters & AUC\\
    \midrule
   Recency Rule&$L=7$&0.700\\
   Frequency Rule&$M=1, D=15$&0.711\\
   Logistic Regression&-&0.844\\
   Distributed Factorization Machine&-&0.846\\
   TCCP&$OP=15$&0.866\\
  \bottomrule
\end{tabular}
\vskip -0.15in
\end{table}

\section{Conclusion}
In this paper, we proposed a time-sensitive customer churn prediction (TCCP) framework, which first collects enough positive samples with better timeliness by shortening observation period, and then applies PU learning technique to obtain a classifier. We applied Distribute Factorization Machines to deal with the massive data. We performed experiments on real industry data from \textsl{Alipay.com}, and the results demonstrated that, compared with the rule-based methods and the traditional supervised learning methods, our model achieves better results.

\bibliographystyle{ACM-Reference-Format}
\bibliography{sigproc} 

\end{document}